\documentclass{article} %
\usepackage{natural_galore_template/natural_galore,times}

\usepackage{amsmath,amsfonts,bm}

\def\eqref#1{equation~\ref{#1}}

\def\1{\bm{1}}

\DeclareMathAlphabet{\mathsfit}{\encodingdefault}{\sfdefault}{m}{sl}
\SetMathAlphabet{\mathsfit}{bold}{\encodingdefault}{\sfdefault}{bx}{n}

\usepackage{xcolor}
\usepackage{fancyhdr,graphicx}
\usepackage{amssymb,amsthm,amsmath}
\usepackage{booktabs,array} %
\usepackage{thmtools}
\usepackage{thm-restate}
\usepackage[title]{appendix}
\usepackage{tikz}
\usepackage{resizegather}
\usepackage{wrapfig}
\usepackage{enumitem,kantlipsum}
\usepackage{mathtools}
\usepackage{nccmath}

 \usepackage{xspace}

\usepackage{dsfont}
\usepackage{stmaryrd}

\renewcommand{\phi}{\varphi}

\usepackage{hyperref}
\usepackage{url}
\usepackage{forest}
\usepackage{tikz}
\usepackage{amsmath}
\usepackage{amssymb}
\usepackage{mathtools}
\usepackage{amsthm}
\usepackage{algpseudocode}
\usepackage{float}
\restylefloat{table}
\usetikzlibrary{shapes.geometric, arrows}

\usepackage[capitalize,noabbrev]{cleveref}
\theoremstyle{plain}

\theoremstyle{definition}

\usepackage{color, colortbl}
\definecolor{LightCyan}{rgb}{0.94,1,1}

\usepackage[textsize=tiny]{todonotes}
\usepackage{multirow}

\usepackage{graphicx}
\usepackage{pifont}
\usepackage{xcolor,colortbl}
\usepackage{multicol}
\usepackage{tablefootnote}
\usepackage[para,online,flushleft]{threeparttable}
\usepackage{xspace}

\usepackage{setspace}

\usepackage[many]{tcolorbox}

\usepackage{subcaption}

\usepackage{letltxmacro}

\usepackage{siunitx}
\usepackage{xcolor}
\usepackage{listings}

\usepackage[ruled,vlined]{algorithm2e}
\definecolor{commentcolor}{RGB}{110,154,155}   %

\tikzstyle{startstop} = [rectangle, rounded corners, minimum width=3cm, minimum height=1cm,text centered, draw=black, fill=red!30]
\tikzstyle{process} = [rectangle, minimum width=3cm, minimum height=1cm, text centered, draw=black, fill=orange!30]
\tikzstyle{arrow} = [thick,->,>=stealth]
\newcommand{\lowrank}{Natural GaLore\xspace}

\title{\lowrank{}: Accelerating GaLore for memory-efficient LLM Training and Fine-tuning}

\author{Arijit Das \\
ERGO Group AG \\
Düsseldorf, Germany \\
\texttt{arijit.das@selfsupervised.de}
}

\arxiv
\begin{document}

\maketitle

\begin{abstract}
    Training LLMs presents significant memory challenges due to growing size of data, weights, and optimizer states. Techniques such as data and model parallelism, gradient checkpointing, and offloading strategies address this issue but are often infeasible due to hardware constraints. To mitigate memory usage, alternative methods like Parameter-Efficient-Fine-Tuning (PEFT) and GaLore approximate weights or optimizer states. PEFT methods, such as LoRA, have gained popularity for fine-tuning LLMs, though they require a full-rank warm start. In contrast, GaLore allows full-parameter learning while being more memory-efficient. This work introduces \textit{\lowrank}, a simple drop in replacement for AdamW, which efficiently applies the inverse Empirical Fisher Information Matrix to low-rank gradients using Woodbury's Identity. We demonstrate that incorporating second-order information speeds up optimization significantly, especially when the iteration budget is limited. Empirical pretraining on 60M, 130M, 350M, and 1.1B parameter Llama models on C4 data demonstrate significantly lower perplexity over GaLore without additional memory overhead. By fine-tuning RoBERTa on the GLUE benchmark using \textit{\lowrank}, we demonstrate significant reduction in gap 86.05\% vs 86.28\% for full-finetuning. Furthermore, fine-tuning the TinyLlama 1.1B model for function calling using the TinyAgent framework shows that \textit{\lowrank} achieving 83.09\% accuracy on the TinyAgent dataset, significantly outperforms 16-bit LoRA at 80.06\% and even surpasses GPT4-Turbo by 4\%, all while using 30\% less memory. \footnote[1]{All code to reproduce the results are available at: https://github.com/selfsupervised-ai/Natural-GaLore.git}
\end{abstract}

\vspace{-4mm}
\section{Introduction}
Large Language Models (LLMs) have achieved remarkable performance across various disciplines, including conversational AI and language translation. However, training and fine-tuning these models demand enormous computational resources and are highly memory-intensive. This substantial memory requirement arises from storing billions of trainable parameters along with associated gradients and optimizer states.

To quantify this, consider a model with $\Psi$ parameters which is being trained using the Adam optimizer. In this case, storing parameters and their gradients in 16-bit precision formats like FP16 or BF16 requires $2\Psi$ bytes each. The associated optimizer states are typically stored in 32-bit precision (FP32) for numerical stability, necessitating an additional $4\Psi$ bytes for each parameter, gradient momentum, and variance, amounting to $12\Psi$ bytes. Therefore, the total memory requirement sums up to $16\Psi$ bytes. When accounting for model-dependent memory, such as activations during forward and backward passes, and residual memory, like temporary buffers and memory fragmentation, the overall memory footprint can easily exceed $18\Psi$ bytes \citep{raffelExploringLimitsTransfer2020,touvronLlamaOpenFoundation2023,chowdheryPaLMScalingLanguage2022}.

This enormous memory demand poses significant challenges, especially when training LLMs on hardware with limited memory capacity. As models continue to scale, efficient memory utilization becomes critical for making training feasible and accessible. In this work, we develop an efficient adaptation to the GaLore algorithm \citep{zhao2024galore}, which significantly reduces the memory footprint during training and fine-tuning of LLMs by approximating the optimizer state. Our approach, \textit{\lowrank}, leverages the low-rank structure of gradients and incorporates second-order information to achieve faster convergence and higher performance without additional memory overhead and can be used as a drop in replacement to standard optimization algorithms like Adam and AdamW.

\paragraph{Parallel and Distributed Training Techniques}
Researchers have developed various distributed computing techniques that leverage system-level optimizations and hardware resources to mitigate the substantial memory requirements in training LLMs.

One prominent framework is \textit{Distributed Data-Parallel (DDP)} that combines data parallelism where the training dataset is partitioned across multiple devices or nodes, with efficient gradient synchronization mechanisms, minimizing communication overhead. While data parallelism efficiently utilizes multiple GPUs, it can still face memory bottlenecks when model sizes exceed the memory capacity of a single device.

\textit{Model parallelism} addresses this limitation by partitioning the model across multiple devices, allowing for the training of models that are too large to fit into the memory of a single GPU. Techniques like \textit{pipeline parallelism} \citep{huangGPipeEfficientTraining2019} and \textit{tensor parallelism} \citep{shoeybiMegatronLMTuningScaling2019} enables the distribution of different layers or partitions of layers across devices. However, model parallelism introduces communication overhead and can be complex to implement effectively.

Another effective technique is \textit{gradient checkpointing} \citep{chenTrainingDeepNets2016}, which reduces memory usage by selectively storing only a subset of activations during the forward pass and recomputing them during the backward pass as needed. This approach trades increased computational overhead for reduced memory consumption, enabling the training of deeper models without exceeding memory constraints.

\textit{Memory offloading} strategies, such as those implemented in ZeRO-Offload \citep{rajbhandariZeROMemoryOptimizations2020}, move optimizer states and gradients to CPU memory when not actively in use, freeing up GPU memory for other operations. ZERO can also partition optimizer states and gradients across DDP processes, eliminating redundancy and significantly reducing memory footprint. \textit{Fully Sharded Data Parallel} \citep{zhaoExtendingTorchElasticStateful2020} extends this concept by sharding model parameters in addition to optimizer states and gradients.

These system-level optimizations have been instrumental in training state-of-the-art LLMs such as LLaMA3 \citep{touvronLlamaOpenFoundation2023}, GPT-3 \citep{brownLanguageModelsAre2020}, Mistral \citep{jiangMistralEfficientComposable2023}, and Gopher \citep{raeScalingLanguageModels2021} on multi-node, multi-GPU clusters.

While these distributed computing solutions enable the training of large models by leveraging extensive hardware resources, they come with increased system complexity and operational costs. Therefore, there is a pressing need for alternative approaches that reduce memory consumption without relying solely on distributed computing resources. Optimization techniques that approximate parameters or optimizer states offer a promising direction for making LLM training more accessible and efficient.

\paragraph{Parameter-Efficient Fine-Tuning}

PEFT techniques efficiently adapt pre-trained language models to various downstream applications without fine-tuning all the model's parameters \citep{dingDeltaTuningComprehensive2022}, significantly reducing the computational and memory overhead.

Among these techniques, the popular LoRA \citep{huLoRALowRankAdaptation2021} parametrizes a weight matrix $W \in \mathbb{R}^{n \times m}$ as:
\begin{equation}
 W = W_0 + BA,
\end{equation}
where $W_0$ is a frozen full-rank pre-trained weight matrix, and $B \in \mathbb{R}^{n \times r}$ and $A \in \mathbb{R}^{r \times m}$ are trainable low-rank adapters to be learned during fine-tuning. Since the rank $r \ll \min(m, n)$, the adapters $B$ and $A$ contain significantly fewer trainable parameters, reducing memory requirements for both parameter and optimizer states.

LoRA has been extensively used to reduce memory usage during fine-tuning, effectively enabling large models to be adapted to new tasks with minimal additional memory overhead. There are a few variants of LoRA proposed to enhance its performance \citep{renduchintalaTiedLoraEnhacingParameter2023, shengSLoRAServingThousands2023, zhangLORAFAMEMORYEFFICIENTLOWRANK, xiaChainLoRAEfficient2024}, supporting multi-task learning \citep{wangMultiLoRADemocratizingLoRA2023}, and further reducing the memory footprint \citep{dettmersQLoRAEfficientFinetuning2023}. Its variant, ReLoRA \citep{lialinReLoRAHighRankTraining2023}, extends LoRA's approach to pre-training by periodically updating the frozen weight matrix $W_0$ using the previously learned low-rank adapters. This incremental updating allows for continual learning without storing entire optimizer states for all parameters, leading to faster training times and lower computational costs. Furthermore, this allows for rapid adaptation of large models to multiple downstream tasks without storing separate copies of the entire model for each task.

Despite their benefits, recent works have highlighted several limitations of low-rank reparameterization approaches. LoRA does not consistently achieve performance comparable to full-rank fine-tuning, particularly in complex tasks \citep{xiaChainLoRAEfficient2024}. In pre-training from scratch, methods like ReLoRA require an initial phase of full-rank model training as a warmup before optimizing in the low-rank subspace \citep{lialinReLoRAHighRankTraining2023}. The shortcomings of low-rank parameter reparameterization suggest that alternative strategies are needed to achieve both memory efficiency and high performance.

\paragraph{Gradient Low-Rank Projection (GaLore)}

An alternative to parameter approximation is the approximation of the optimizer states. By reducing the memory footprint associated with optimizer states, it is possible to maintain full-parameter learning—thus preserving model capacity and performance—while achieving significant memory savings.

The core idea behind GaLore \citep{zhao2024galore} is to exploit the slowly changing low-rank structure of the gradient matrix $g \in \mathbb{R}^{n \times m}$, rather than approximating the weights. During neural network training, gradients naturally exhibit low-rank properties, a phenomenon studied extensively in both theoretical and practical settings \citep{zhaoZerOInitializationInitializing2022,cossonLowRankGradientDescent2023,yang2023spectral}. This intrinsic low-rank structure of gradients has been applied to reduce communication costs \citep{wangATOMOCommunicationefficientLearning,vogelsPowerGossipPracticalLowRank2020} and to decrease memory footprints during training \citep{gooneratneLowrankGradientApproximation2020,huangLowRankGradientDescent2023}.

Specifically, consider the compact SVD decomposition of the gradient matrix \(\mathbf{g} = \mathbf{P} \Sigma \mathbf{Q}^{T}\), where \(\mathbf{P} \in \mathbb{R}^{n \times r}\) and \(\mathbf{Q} \in \mathbb{R}^{m \times r}\) are the associated semi-orthognal matrices.  Then, GaLore projects the gradient matrix $\mathbf{g}$ into a low-rank form:
\begin{equation}
    \mathbf{g}_{\text{low-rank}} = \mathbf{P}^{T} \mathbf{g}.
\end{equation}
Here, $r \ll \min(n, m)$ is the target rank, \(n\) is the parameter count, \(m\) is the batch size and $\mathbf{g}_{\text{low-rank}}$ serves as an efficient approximation of the original gradient. The projection matrix $\mathbf{P}$ is updated periodically (e.g., every 200 iterations), which incurs minimal amortized computational cost.

By operating on low-rank approximations of the gradients, GaLore significantly reduces the memory footprint, leading to up to \textbf{30\%} memory reduction compared LoRA \citep{zhao2024galore}. Moreover, GaLore maintains full-parameter learning, allowing updates to all model parameters, leading to better generalization and performance than low-rank adaptation methods. Further, GaLore is agnostic to the choice of optimizer and can be easily integrated into existing optimization algorithms with minimal code modifications.

While GaLore offers significant memory savings and enables full-parameter learning, its performance has yet to match that of optimizers in full optimizer state space. Reliance on low-rank gradient approximations may not fully capture the rich optimization dynamics. These limitations suggest that while GaLore is a valuable step toward memory-efficient training, further enhancements are necessary to bridge the performance gap with standard optimizers.

\paragraph{Our Approach}

In this work, we propose to bridge the gap by incorporating a second-order regularizer into the low-rank gradient estimate, which adjusts parameter updates more effectively, leading to faster convergence. We show that applying the inverse of the empirical Fisher Information Matrix (FIM) to the low-rank gradients leads to variance reduction of the gradient estimate, incorporates information about the curvature of the loss landscape, and reduces dependence on the starting point. All of these lead to significantly faster convergence, especially in a limited iteration regime.

We introduce the \textit{\lowrank} algorithm, a matrix-free algorithm for efficiently applying the inverse FIM to the low-rank gradients, using Woodbury Identity, Cholesky Decomposition, and Matrix-Vector Products, all of which can be efficiently implemented on the GPU. Further, our approach does not require any explicit layer-wise information or significant computational overhead, as is seen in existing approaches like K-Fac \citep{martens2015optimizing}.

 We validate the effectiveness of \textit{\lowrank} through extensive empirical evaluations. Pre-training experiments on LLaMA models with 60M, 300M, and 1.1B parameters using the C4 dataset demonstrate that \textit{\lowrank} achieves significantly lower perplexity than GaLore without additional memory overhead, indicating faster convergence within the same computational budget.

 Furthermore, we showcase the practical benefits of \textit{\lowrank} in fine-tuning tasks. We fine-tune the TinyLlama 1.1B model for function calling using the TinyAgent framework. Our results show that \textit{\lowrank} significantly outperforms LoRA in this setting, achieving an accuracy of \textbf{83.09\%} on the TinyAgent dataset. This performance significantly surpasses 16-bit LoRA and exceeds that of GPT-4-turbo by 4\%, all while using \textbf{30\%} less memory.

\section{Accelerating GaLore with Natural Gradients}

\subsection{Next Token Prediction}

Generative LLMs are trained to predict the next token in a sequence based solely on the previously observed tokens. This "causal" approach respects the temporal order of language, ensuring that the model's predictions at any point depend only on past and not future inputs.

Given a sequence of tokens \( x = (x_1, x_2, \dots, x_T) \), the objective is to maximize the likelihood of a sequence by decomposing it into a product of conditional probabilities:

\begin{eqnarray}
\text{Prob}_{\mathbf{\theta}}(x) = \prod_{t=1}^T \text{Prob}_{\mathbf{\theta}}(x_t \mid x_{<t})
\end{eqnarray}

where \( x_{<t} = (x_1, x_2, \dots, x_{t-1}) \) represents all tokens before position \( t \) and \( \text{Prob}_{\mathbf{\theta}}(x_t \mid x_{<t}) \) is the probability of the next token given all previous tokens and the parameter \( \mathbf{\theta} \in \mathbb{R}^{n \times m} \).

This is equivalent to minimizing the Negative Log-Likelihood (NLL) of the observed sequences, which is the cross-entropy loss between the predicted probability distribution and the actual next token:

\begin{eqnarray}
\Phi(\mathbf{\theta}) = -\sum_{t=1}^T \log \text{Prob}_{\mathbf{\theta}}(x_t \mid x_{<t})
\label{eq:cross_entropy_loss}
\end{eqnarray}

This loss penalizes the model more when it assigns lower probabilities to the correct next token. By minimizing this loss, the model learns to assign higher probabilities to appropriate continuations of text. However, the loss is non-convex and high-dimensional, for LLMs the dataset is also massive, making the optimization problem very challenging.

\subsection{Low-Rank Gradient Descent}

Stochastic gradient descent algorithms are iterative, where each step aims to find the optimal update direction that minimizes the loss function locally. Now in the case of GaLore, the update direction is restricted to the affine subspace \(\mathbf{u}_{k} \in {\mathbf{\theta}_{k}} + \text{Range} \left(\mathbf{P}_{k}\right)\). Here \(\mathbf{P}_{k} \in \mathbb{R}^{n \times r}\) is the left projection matrix, calculated using the compact SVD decomposition of the gradient matrix \(\nabla_{\mathbf{\theta}} \Phi(\mathbf{\theta}_{k}) = \mathbf{P}_{k} \Sigma \mathbf{Q}_{k}^{T}\).

Then, the local neighborhood around this update can be defined using the Taylor series expansion \citep{lin2022randomized}:

\begin{eqnarray}
\Phi(\mathbf{\theta}_{k} + \mathbf{P}_{k} \mathbf{u}_{k}) \approx \Phi(\mathbf{\theta}_{k}) + \mathbf{g}_{k}^{T}\mathbf{u}_{k} + \frac{1}{2} \mathbf{u}_{k}^{T} \mathbf{H}_{k}  \mathbf{u}_{k}
\label{eq:taylor_series_expansion}
\end{eqnarray}

where \(\mathbf{g}_{k} = \mathbf{P}_{k}^{T}\nabla_{\mathbf{\theta}} \Phi(\mathbf{\theta}_{k})\) is the low rank projected gradient and \(\mathbf{H_{k}} = \mathbf{P}_{k}^{T}\nabla^2_{\mathbf{\theta}} \Phi(\mathbf{\theta}) \mathbf{P}_{k}\) is the Hessian matrix.

However, the Hessian matrix \(\mathbf{H}_{k}\) is often computationally expensive to compute and store, especially for large-scale language models (LLMs) with billions of parameters. Fortunately, precisely under the condition that the loss function can be represented in terms of KL divergence between the actual and approximated distributions [\ref{eq:cross_entropy_loss}], then \(\mathbf{H_{k}}\) can be approximated by the FIM. The FIM is defined as the expectation of the Hessian of the negative log-likelihood w.r.t. the data distribution:

\begin{eqnarray}
\mathbf{F}_{k} = \mathbb{E}_{x \sim p_{\text{data}}} \left[ \mathbf{H}_{k} \right]
\end{eqnarray}

The FIM captures the curvature of the loss landscape and provides a natural metric for the optimization process. Hence, it can better adjust parameter updates according to the geometry of the parameter space. However, as the theoretical data distribution is unknown, in practice, we need to estimate it using the empirical FIM \citep{martensNewPerspectiveNatural2014} defined by:

\begin{eqnarray}
\mathbf{\hat{F}}_{k} = \frac{1}{h} \sum_{k=1}^{h} \mathbf{g_{k}} \mathbf{g_{k}}^{T}
\end{eqnarray}

where \(h\) is the history of gradients from past batches we would like to consider. Then, the optimal direction \(\mathbf{u}_{k}^{*}\), which minimizes the loss in this local neighborhood, is given by (cite Fuji et al. paper):

\begin{eqnarray}
\mathbf{u}_{k}^{*} &=& \mathbf{\hat{F}}_{k}^{-1} \mathbf{g}_{k}
\label{eq:optimal_direction}
\end{eqnarray}

This leads to the optimal gradient descent update step:

\begin{eqnarray}
\mathbf{\theta}_{k+1} = \mathbf{\theta}_{k} - \eta \mathbf{P}_{k} \mathbf{u}_{k}^{*}
\label{eq:gradient_descent_update}
\end{eqnarray}

for some learning rate \(\eta\).

Many popular stochastic optimization algorithms approximate the diagonal of the empirical FIM using second-moment estimates of the gradient \(\mathbf{g}_{k}\), which when added with Polyak style parameter averaging (i.e., momentum), asymptotically achieve the optimal Fisher efficient convergence rate \citep{martens2020new}.

For instance, in the case of Adam \citep{kingmaAdamMethodStochastic2014}, the optimal update step is approximated by including the momentum term \(\mathbf{m}_{k} \in \mathbb{R}^{r\times m}\) and the learning rate \(\eta\) is scaled by the square root of the second moment estimate \(\mathbf{v}_{k} \in \mathbb{R}^{r\times m}\). With all operations being elementwise, the update direction becomes:

\begin{eqnarray}
    \mathbf{m}_{k} &=& \beta_{1} \mathbf{m}_{k-1} + (1-\beta_{1}) \mathbf{g}_{k} \\
    \mathbf{v}_{k} &=& \beta_{2} \mathbf{v}_{k-1} + (1-\beta_2) \mathbf{g}^{2}_{k}  \\
    \mathbf{u_{k}^{*}} &=& \mathbf{m}_{k} / \sqrt{\mathbf{v}_{k} + \epsilon}
    \label{eq:adam_update}
\end{eqnarray}

This update, when applied to [\ref{eq:gradient_descent_update}], gives the GaLore optimization algorithm, which is memory efficient as it only requires storing the projection matrix and the costly optimizer states \(\left(g_{k}, m_{k}, v_{k}\right)\) are now significantly reduced by a factor of \(\frac{n}{r}\), where the rank \(r\), can be chosen based on the tradeoff between memory limitations and performance requirements.

\subsection{\lowrank and Fisher Efficiency}

Despite clear advantages, the performance of GaLore is not on par with AdamW \citep{loshchilov2017decoupled} optimization on the original space. To bridge this gap, we propose \textit{\lowrank}, which uses the full empirical FIM, thereby incorporating the missing second-order interaction information in the optimization process.

As we now argue, this leads to a much more favorable dependence on the starting point, which means that the optimizer can make much more progress given a limited iteration budget. Further, when using a decaying learning rate schedule like with AdamW \citep{loshchilov2017decoupled}, the asymptotic convergence rate can be faster \citep{martens2020new} by a significantly large constant factor.

Natural gradient descent is known \citep{martens2020new} to be Fisher efficient, precisely for our loss function [\ref{eq:cross_entropy_loss}]. Fisher efficiency means that the natural gradient estimator asymptotically achieves the lowest possible variance among all unbiased gradient estimators.

For \textit{\lowrank}, the gradient descent update [\ref{eq:gradient_descent_update}] leads to a sequence of estimates \( \mathbf{\theta}_{k} \) whose variance satisfies \citep{amariNaturalGradientWorks1998}:

\begin{eqnarray}
\text{Var}[\mathbf{\theta}_{k}] = \frac{1}{mk} \mathbf{F}_{k}^{-1}(\mathbf{\theta}_{k}^*) + \mathcal{O}\left(\frac{1}{k^2}\right)
\label{eq:variance_reduction}
\end{eqnarray}

which is asymptotically the smallest possible variance matrix satisfying the Cramér-Rao lower bound, that any unbiased estimator computed from \(mk\) training samples can have, with \(m\) being the batch size.

Here, \(\mathbf{\theta}_{k}^*\) is the local optimum in the neighborhood defined by the Taylor series expansion [\ref{eq:taylor_series_expansion}] around the update direction. This is an important caveat, as the guarantee is only for local convergence in a convex neighborhood. The loss function is non-convex, so the property can not be stated to hold for the global optimum.

The result also relies on the computation of the exact FIM \( \mathbf{F}_{k}(\mathbf{\theta}_{k}) \) using the entire data distribution, which is not practical. The Fisher efficiency guarantee is, however, only approximately satisfied when using the empirical FIM \(\mathbf{\hat{F}_{k}}\) instead. Nevertheless, we still get a variance reduction in the gradient estimates, leading to faster convergence and better optimization performance in the early stages of training large-scale models, making it especially valuable for training with a limited iteration budget.

Further, incorporating second-order information through the empirical FIM allows the optimizer to account for the curvature of the loss landscape, enabling natural gradient descent to take more informed steps than standard gradient descent, potentially escaping flat regions or navigating steep ravines more effectively.

In \citep{martens2020new}, it was shown that the expected update direction can be expressed as a sum of two terms, one that scales as \(\mathcal{O}(1/k)\), which is independent of the starting point and another that scales as \(\mathcal{O}(1/k^2)\), which is dependent on the starting point. If momentum is applied to the gradient estimator, the first term becomes independent of the choice of FIM estimator, thereby not leading to any asymptotic improvements. However, regularizing with the empirical FIM estimate can significantly reduce the constant factor associated with the starting-point-dependent second term. This leads to practical performance gains in finite iteration regimes (although negligible for large \(k\)).

Finally, the Fisher efficiency result also assumes that the model can perfectly capture the data distribution, a condition known as \emph{realizability}. However, with the growing size of LLMs, this assumption is likely to hold, thereby satisfying the conditions for the guarantee. Therefore, especially in low-resource settings, \textit{\lowrank} can be a promising approach for training LLMs under memory constraints.

\subsection{Natural Gradient Transform}

Our \textit{\lowrank} algorithm is designed to efficiently apply the inverse empirical FIM to low-rank gradients using Woodbury's Identity. Most of the steps in the algorithm are similar to GaLore \citep{zhao2024galore}, with the critical difference being the incorporation of the natural gradient transform.

In order to implement the natural gradient transform, we compute the inverse of the empirical FIM and apply it to the gradient \(\mathbf{g_{k}}\) using Woodbury's Identity, which allows us to efficiently compute the inverse of a matrix of the form \(A + UBU^T\). Woodbury's Identity states that:

\begin{eqnarray}
(A + UBU^T)^{-1} = A^{-1} - A^{-1}U(B^{-1} + U^TA^{-1}U)^{-1}U^TA^{-1}
\end{eqnarray}

Now, if we choose \(\mathbf{\hat{F}}_{k} = \lambda I + GG^{T}\), \(A = \lambda I\), \(U = G\), and \(B = I\), where \(G = [\operatorname{vec}(\mathbf{g}_{k}), \operatorname{vec}(\mathbf{g}_{k-1}),\ldots, \operatorname{vec}(\mathbf{g}_{k-s})]\) is the stacked gradient matrix over the past \(s\) gradients and \(\lambda\) is a small constant for Tikhonov regularization, then, the inverse of the empirical FIM applied to the gradient \(\mathbf{g_{k}}\) i.e. the natural gradient \(\mathbf{\tilde{g}}_{k} = \mathbf{\hat{F}}_{k}^{-1}\mathbf{g_{k}}\) can be calculated as:

\begin{eqnarray}
 \mathbf{\tilde{g}}_{k} = \frac{1}{\lambda}\mathbf{g_{k}} - \frac{1}{\lambda}G\left(\lambda I + G^{T}G\right)^{-1}G^{T}\mathbf{g_{k}}
\end{eqnarray}

To compute the above formula efficiently, let \(S = I + \frac{1}{\lambda}G^{T}G \in \mathbb{R}^{s\times s}\) and \(y = G^T\mathbf{g_{k}}\). Cholesky decomposition is used to solve for \(z\) in
\begin{eqnarray}
S z = y
\end{eqnarray}
which requires only \(\mathcal{O}(s^2)\) time. Then, the final natural gradient estimate can be computed using only matrix-vector products, which is very memory efficient:
\begin{eqnarray}
 \mathbf{\tilde{g}}_{k} = \frac{1}{\lambda}\mathbf{g_{k}} - \frac{1}{\lambda^{2}}Gz
\end{eqnarray}
This natural gradient estimate \(\mathbf{\tilde{g}}_{k}\) can then be sent to the Adam optimizer [\ref{eq:adam_update}], and the model parameters the same way as in GaLore.

\section{Experiments}

We evaluate \textit{\lowrank} on pre-training and fine-tuning tasks for LLMs. All experiments are conducted on a single node with 8 NVIDIA A100 GPUs to leverage high-performance computing capabilities, yet stay within reasonable limits.

\subsection{Pre-training on the C4 Dataset}

To assess the effectiveness of \textit{\lowrank}, we apply it to pre-train LLaMA-based language models of sizes ranging from 60 million to 1.1 billion parameters, on the C4 dataset. The C4 dataset is a colossal, cleaned version of the Common Crawl Corpus, primarily intended for pre-training language models and word representations \citep{raffelExploringLimitsTransfer2020}. It provides a diverse and extensive corpus, making it suitable for evaluating pre-training methods in realistic scenarios.

We adopt the experimental setup from \citet{lialinReLoRAHighRankTraining2023}, utilizing a LLaMA-based\footnote[2]{LLaMA materials in our paper are subject to the LLaMA community license.} architecture with RMSNorm and SwiGLU activations \citep{shazeerGLUVariantsImprove2020,touvronLlamaOpenFoundation2023}. We maintain the same set of hyperparameters for each model size across all methods, except for the learning rate, which is tuned individually to ensure optimal performance. All experiments use the BF16 format to reduce memory usage without compromising computational efficiency, the same computational budget and the best validation perplexity is reported.

\begin{figure}[ht]
    \centering
    \vspace{-2.5mm}
    \includegraphics[width=\linewidth]{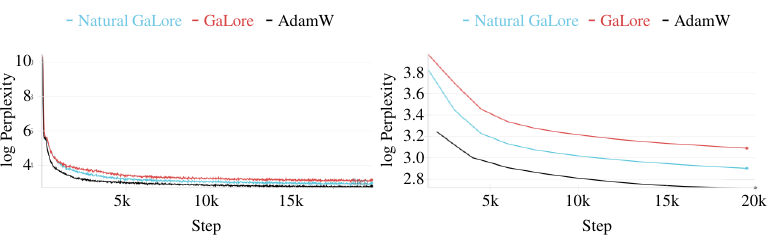}
    \vspace{-5.5mm}
    \caption{\small{Training and Validation log Perplexity for Llama 1.1B
    }}
    \vspace{-5mm}
    \label{fig:train_loss}
\end{figure}

\begin{table*}[ht]
    \centering
    \caption{\small{Comparison of \textit{\lowrank} with other low-rank algorithms on pre-training various sizes of LLaMA models on the C4 dataset. Validation log perplexity is reported (averaged over 5 runs), along with a memory estimate (in gigabytes) of the total parameters and optimizer states based on BF16 format.}}
    \label{tab:lora_compare_llama}
    \begin{tabular}{lcccc}
    \toprule
                     & \textbf{60M} & \textbf{130M} & \textbf{350M} & \textbf{1.1B} \\
    \midrule
    Full-Rank        & 3.52 (0.36G) & 3.22 (0.76G) & 2.93 (2.06G) & 2.72 (7.80G) \\
    \midrule
    \textit{\lowrank} & \textbf{3.53} (0.24G) & \textbf{3.22} (0.52G) & \textbf{2.93} (1.22G) & \textbf{2.80} (4.38G) \\
    GaLore           & 3.56 (0.24G) & 3.24 (0.52G) & 2.95 (1.22G) & 2.90 (4.38G) \\
    Low-Rank         & 4.35 (0.26G) & 3.82 (0.54G) & 3.62 (1.08G) & 4.96 (3.57G) \\
    LoRA             & 3.55 (0.36G) & 3.52 (0.80G) & 3.24 (1.76G) & 2.96 (6.17G) \\
    ReLoRA           & 3.61 (0.36G) & 3.38 (0.80G) & 3.37 (1.76G) & 2.91 (6.17G) \\
    \bottomrule
    Rank $r / d_{\text{model}}$ & 128 / 256 & 256 / 768 & 256 / 1024 & 512 / 2048 \\
    Training Tokens  & 1.1B & 2.2B & 6.4B & 13.1B \\
    \bottomrule
    \end{tabular}
\end{table*}

Table~\ref{tab:lora_compare_llama} presents the validation perplexity and memory consumption for models trained with different methods and Figure~\ref{fig:train_loss} shows the training run for the Llama 1.1B model. Our proposed \textit{\lowrank} consistently outperforms GaLore \citep{zhao2024galore} across all model sizes, achieving validation perplexities closer to the full-rank baseline while maintaining significant memory savings. Furthermore, \textit{\lowrank} exhibits lower perplexities and greater memory consumption compared to other low-rank adaptation methods like LoRA and ReLoRA, due to their less efficient use of low-rank structures and the need for additional optimizer states.

\vspace{-2mm}

\subsection{Fine-Tuning RoBERTa-Base on the GLUE Benchmark}

To further evaluate the effectiveness of \textit{\lowrank}, we conduct experiments on the General Language Understanding Evaluation (GLUE) benchmark using the pre-trained RoBERTa-Base model. The GLUE benchmark is a collection of nine natural language understanding tasks, including single-sentence tasks like CoLA \citep{warstadt2019neural}, similarity and paraphrase tasks like MRPC \citep{dolan2005automatically} and STS-B \citep{cer2017semeval}, and inference tasks like RTE \citep{dagan2005pascal}, MNLI \citep{williams2018broad}, and QNLI \citep{rajpurkar2016squad}. This benchmark is widely used to assess the performance of language models on diverse linguistic phenomena.

In our experiments, we fine-tune the RoBERTa-Base model using \textit{\lowrank} and compare its performance with full fine-tuning and LoRA \citep{huLoRALowRankAdaptation2021}. We focus on memory-efficient fine-tuning methods to reduce the computational footprint while maintaining high performance. For each method, we report the average score across all GLUE tasks and individual task scores.

We use the same training hyperparameters across all methods for a fair comparison. The batch size is 32, and we fine-tuned each model for three epochs. The learning rate is selected from \{1e-5, 2e-5, 3e-5\} based on the best validation performance for each task. For \textit{\lowrank} and LoRA, we experiment with rank values of 4 and 8 to study the trade-off between performance and memory efficiency.

Table~\ref{tab:fine_tuning} presents the results of our experiments. \textit{\lowrank} consistently achieves comparable or better performance than LoRA across most tasks while using less memory. Precisely, with a rank of 4, \textit{\lowrank} attains an average score of \textbf{86.05}, closely matching the complete fine-tuning baseline of 86.28 and outperforming LoRA's average score of 85.61. This demonstrates that \textit{\lowrank} can effectively fine-tune large models with reduced memory consumption without sacrificing performance.

\begin{table}[ht]
    \caption{\small{Evaluating \textit{\lowrank} for memory-efficient fine-tuning on the GLUE benchmark using pre-trained RoBERTa-Base. We report the average score of all tasks. Memory consumption is reported in millions of parameters (M).}}
    \label{tab:fine_tuning}
    \centering
    \resizebox{\linewidth}{!}{%
    \begin{tabular}{l|c|cccccccc|c}
    \toprule
               & \textbf{Memory} & \textbf{CoLA} & \textbf{STS-B} & \textbf{MRPC} & \textbf{RTE} & \textbf{SST-2} & \textbf{MNLI} & \textbf{QNLI} & \textbf{QQP} & \textbf{Avg} \\
    \midrule
    Full Fine-Tuning & 747M & 62.24 & 90.92 & 91.30 & 79.42 & 94.57 & 87.18 & 92.33 & 92.28 & 86.28 \\
    \midrule
    \textbf{\textit{\lowrank} (rank=4)} & 253M & 61.50 & \textbf{90.80} & \textbf{92.10} & \textbf{79.50} & \textbf{94.20} & \textbf{87.05} & \textbf{92.30} & 91.15 & \textbf{86.05} \\
    GaLore (rank=4) & 253M & 60.35 & 90.73 & 92.25 & 79.42 & 94.04 & 87.00 & 92.24 & 91.06 & 85.89 \\
    LoRA (rank=4) & 257M & \textbf{61.38} & 90.57 & 91.07 & 78.70  & 92.89 & 86.82 & 92.18 & \textbf{91.29} & 85.61 \\
    \midrule
    \textbf{\textit{\lowrank} (rank=8)} & 257M & 61.70 & \textbf{90.90} & \textbf{92.25} & \textbf{79.80} & \textbf{94.40} & \textbf{87.20} & \textbf{92.35} & \textbf{91.25} & \textbf{86.23} \\
    GaLore (rank=8) & 257M & 60.06 & 90.82 & 92.01 & 79.78 & 94.38 & 87.17 & 92.20 & 91.11 & 85.94 \\
    LoRA (rank=8) & 264M & \textbf{61.83} & 90.80 & 91.90 & 79.06  & 93.46 & 86.94 & 92.25 & 91.22 & 85.93 \\
    \bottomrule
    \end{tabular}
    }
    \vskip -0.1in
\end{table}

\vspace{-2mm}

\subsection{Fine-Tuning TinyLlama 1.1B for Function Calling in Advanced Agentic Systems}

Advanced Agentic Systems (AAS) require language models that can understand and generate code snippets to integrate various tools and APIs, fulfilling user queries through function-calling. We utilize the TinyAgent framework, which provides an end-to-end pipeline for training and deploying task-specific LLM agents capable of efficient and accurate function-calling \citep{erdogan2024tinyagent} to drive agentic systems at the edge.

Given a natural language query, the LLM agent must generate a sequence of pre-defined function-calls that accomplish the desired tasks. The challenge lies in determining the appropriate arguments, to call the correct functions, in the right order while respecting interdependencies among the functions.

LLMCompiler \citet{kim2023llmcompiler}, is a framework that enables language models to perform function-calling by first generating a function-calling plan, which includes the required functions and arguments. The LLMCompiler then compiles this plan into an executable sequence of function-calls. The critical aspect is training the model to produce a function-calling plan with the correct syntax and dependencies.

The off-the-shelf pre-trained TinyLlama 1.1B (instruct-32k) model performs poorly on this task. The model generates incorrect sets of functions, hallucinated function names, fails to respect dependencies, and passes arguments incorrectly. This underperformance is expected, as the model was initially trained on datasets like SlimPajama and StarCoder, which are not specific to function-calling tasks. To address this, we follow the TinyAgent framework \citep{erdogan2024tinyagent} and fine-tune the TinyLlama 1.1B model on a high-quality, curated dataset designed for function-calling.

\paragraph{TinyAgent Dataset}

The TinyAgent dataset \citep{erdogan2024tinyagent} is a meticulously curated collection aimed at building a local agentic system for function-calling on Apple MacBooks for day-to-day tasks. It contains 40K examples of natural language queries and corresponding function-calling plans. The dataset is divided into 38K training examples, 1K validation examples, and 1K test examples. It encompasses 16 tasks, including Email, Contacts, SMS, Calendar, Notes, Reminders, File Management and Zoom Meetings. Each task has predefined scripts that the model needs to generate. The dataset is intentionally challenging, requiring the model to understand dependencies between function-calls and the arguments to be passed.

\paragraph{Fine-Tuning Procedure}

We fine-tune the TinyLlama 1.1B model on the TinyAgent dataset for three epochs using a batch size of 32. The learning rate is set to \(7 \times 10^{-5}\). After each epoch, the model is evaluated on the validation set, and the best-performing model is selected based on validation performance to be evaluated on the test set.

During fine-tuning, the prompt includes descriptions of the ground truth functions and irrelevant functions serving as negative samples. This strategy encourages the model to learn to select the correct functions rather than merely memorizing the ground truth. Additionally, several in-context examples demonstrate how queries are translated into function-calling plans. These examples are selected using a Retrieval-Augmented Generation (RAG) process based on the user's query from the training data and a DeBERTa-v3-small model \citep{he2021debertav3} fine-tuned for multi-label classification for retrieval among the 16 tools.

The training objective is then to maximize the accuracy of the generated function-calling plans. Success is defined by the model generating the correct plan with the proper set of function-calls, correct arguments, and the appropriate order of function-calls. Verifying the selection of the correct set of functions involves straightforward set comparison. However, ensuring the correctness of arguments and the order of function-calls is more complex and requires constructing the associated Directed Acyclic Graph to check for equality.

\paragraph{Results and Discussion}

\begin{table*}[ht]
\vspace{-3mm}
\caption{
Latency, size, and success rate of TinyAgent models before and after quantization. Latency is the end-to-end latency of the function calling planner, including the prompt processing time and generation.}
\vspace{-1mm}
\begin{center}
\small{
\setlength{\tabcolsep}{6pt}{
\begin{tabular}{c|c|c|c|c}
\toprule
Model &	Weight Precision &	Latency (seconds)	& Model Size (GB)	& Success Rate (\%) \\
\midrule
GPT-3.5 & Unknown & 3.2 & Unknown & 65.04 \\
GPT-4-Turbo & Unknown & 3.9 & Unknown & 79.08 \\
\midrule
\multirow{2}{*}{TinyAgent-1.1B} & 16-bit (\textit{\lowrank}) & 3.9 & 2.2 & \textbf{83.09} \\
& 16-bit (LoRA) & 3.9 & 2.2 & 80.06 \\
\midrule
\multirow{1}{*}{TinyAgent-7B} & 16-bit \citep{erdogan2024tinyagent} & 19.5 & 14.5 & 84.95 \\
\bottomrule
\end{tabular}
}
}
\end{center}
\label{table:t2}
\end{table*}

After fine-tuning, the TinyLlama 1.1B model's success rate on the test set improved significantly. Table~\ref{table:t2} presents the latency, model size, and success rate of various models on the TinyAgent dataset. As shown, \textit{\lowrank} improves the success rate of the 1.1B model from 80.06\% (16-bit LoRA) to \textbf{83.09\%}, also surpassing GPT-4-Turbo by 4\% and approaching the performance of the larger TinyAgent-7B model, which achieves 84.95\%.

These results demonstrate that \textit{\lowrank} not only enhances the performance of smaller models like the 1.1B parameter TinyLlama but also makes them competitive with significantly larger models. By efficiently incorporating second-order information through low-rank natural gradient updates, \textit{\lowrank} enables smaller models to achieve higher accuracy without additional memory overhead.

\section{Conclusion}

We have introduced \textit{\lowrank}, a memory-efficient pre-training and fine-tuning strategy for large language models. \textit{\lowrank} significantly reduces memory usage—by up to 65.5\% in optimizer states—while maintaining or even improving performance in large-scale LLM pre-training and fine-tuning tasks. By incorporating second-order information through an efficient approximation of the inverse Empirical Fisher Information Matrix, \textit{\lowrank} enhances convergence rates, especially in regimes with a limited iteration budget.

Importantly, \textit{\lowrank} can serve as a \emph{drop-in replacement} for standard optimizers like AdamW and integrates seamlessly into existing training pipelines. Our experimental results highlight the \textit{reproducibility} and effectiveness of \textit{\lowrank} across various tasks, including pre-training LLaMA models and fine-tuning on the GLUE benchmark, as well as the TinyAgent function calling tasks. This makes it a compelling choice for large-scale pre-training scenarios where both memory efficiency and model performance are critical.

In the future we want to explore (1) further enhancing memory efficiency by employing low-memory and structured projection matrices, and (2) more extensive empirical evaluation on fine-tuning AAS on a wide variety of tasks. We also hope that our work will inspire future research on memory-efficient training methods from the perspective of optimizer state approximation. We believe that \textit{\lowrank} will be a valuable tool for the community, enabling the training of large-scale models on consumer-grade hardware with limited resources.

\section*{Impact Statement}

This work aims to improve the memory efficiency of training LLMs, thereby reducing the environmental impact of LLM pre-training and fine-tuning. By enabling the training of larger models on hardware with lower memory requirements, our approach helps to minimize energy consumption and carbon footprint associated with training LLMs. Furthermore, by making advanced model training more accessible, we contribute to democratizing AI research and development, allowing a broader community to engage with large-scale models without the need for expensive computational resources.

\bibliography{natural_galore_template/natural_galore}

\begin{thebibliography}{45}
\providecommand{\natexlab}[1]{#1}
\providecommand{\url}[1]{\texttt{#1}}
\expandafter\ifx\csname urlstyle\endcsname\relax
  \providecommand{\doi}[1]{doi: #1}\else
  \providecommand{\doi}{doi: \begingroup \urlstyle{rm}\Url}\fi

\bibitem[Amari(1998)]{amariNaturalGradientWorks1998}
Shun-ichi Amari.
\newblock Natural gradient works efficiently in learning.
\newblock \emph{Neural Computation}, 1998.

\bibitem[Brown et~al.(2020)Brown, Mann, Ryder, Subbiah, Kaplan, Dhariwal,
  Neelakantan, Shyam, Sastry, Askell, et~al.]{brownLanguageModelsAre2020}
Tom~B Brown, Benjamin Mann, Nick Ryder, Melanie Subbiah, Jared Kaplan, Prafulla
  Dhariwal, Arvind Neelakantan, Pranav Shyam, Girish Sastry, Amanda Askell,
  et~al.
\newblock Language models are few-shot learners.
\newblock In \emph{Advances in Neural Information Processing Systems}, 2020.

\bibitem[Cer et~al.(2017)Cer, Diab, Agirre, Lopez-Gazpio, and
  Specia]{cer2017semeval}
Daniel Cer, Mona Diab, Eneko Agirre, I{\~n}igo Lopez-Gazpio, and Lucia Specia.
\newblock Semeval-2017 task 1: Semantic textual similarity multilingual and
  crosslingual focused evaluation.
\newblock In \emph{Proceedings of the 11th International Workshop on Semantic
  Evaluation (SemEval-2017)}, 2017.

\bibitem[Chen et~al.(2016)Chen, Xu, Zhang, and
  Guestrin]{chenTrainingDeepNets2016}
Tianqi Chen, Bing Xu, Chiyuan Zhang, and Carlos Guestrin.
\newblock Training deep nets with sublinear memory cost.
\newblock In \emph{Proceedings of the 20th International Conference on Machine
  Learning (ICML)}, 2016.

\bibitem[Chowdhery et~al.(2022)Chowdhery, Narang, Devlin, Bosma, Mishra,
  Roberts, Barham, Chung, Sutton, Gehrmann,
  et~al.]{chowdheryPaLMScalingLanguage2022}
Aakanksha Chowdhery, Sharan Narang, Jacob Devlin, Maarten Bosma, Gaurav Mishra,
  Adam Roberts, Paul Barham, Hyung~Won Chung, Charles Sutton, Sebastian
  Gehrmann, et~al.
\newblock {PaLM}: Scaling language modeling with pathways.
\newblock \emph{arXiv preprint arXiv:2204.02311}, 2022.

\bibitem[Cosson et~al.(2023)Cosson, Lecouat, Varre, d'Ascoli, and
  Biroli]{cossonLowRankGradientDescent2023}
Victor Cosson, Baptiste Lecouat, Arthur Varre, St{\'e}phane d'Ascoli, and
  Giulio Biroli.
\newblock Low-rank gradient descent converges and generalizes.
\newblock \emph{arXiv preprint arXiv:2301.12995}, 2023.

\bibitem[Dagan et~al.(2006)Dagan, Glickman, and Magnini]{dagan2005pascal}
Ido Dagan, Oren Glickman, and Bernardo Magnini.
\newblock The {PASCAL} recognising textual entailment challenge.
\newblock In \emph{Proceedings of the First International Conference on Machine
  Learning Challenges: Evaluating Predictive Uncertainty, Visual Object
  Classification, and Recognising Textual Entailment}. Springer, 2006.

\bibitem[Dettmers et~al.(2023)Dettmers, Pagnoni, Holtzman, and
  Zettlemoyer]{dettmersQLoRAEfficientFinetuning2023}
Tim Dettmers, Artidoro Pagnoni, Ari Holtzman, and Luke Zettlemoyer.
\newblock {QLoRA}: Efficient finetuning of quantized {LLMs}.
\newblock \emph{arXiv preprint arXiv:2305.14314}, 2023.

\bibitem[Ding et~al.(2022)Ding, Zheng, Wang, Chen, Liu, Zheng, Qiu, Shen, Ding,
  and Tang]{dingDeltaTuningComprehensive2022}
Ning Ding, Xiang Zheng, Yujia Wang, Yifei Chen, Yichi Liu, Haitao Zheng, Xipeng
  Qiu, Yujun Shen, Bolin Ding, and Jie Tang.
\newblock Delta tuning: A comprehensive study of parameter efficient methods
  for pre-trained language models.
\newblock In \emph{Advances in Neural Information Processing Systems}, 2022.

\bibitem[Dolan \& Brockett(2005)Dolan and Brockett]{dolan2005automatically}
William~B Dolan and Chris Brockett.
\newblock Automatically constructing a corpus of sentential paraphrases.
\newblock In \emph{Proceedings of the Third International Workshop on
  Paraphrasing (IWP2005)}, 2005.

\bibitem[Erdogan et~al.(2024)Erdogan, Lee, Jha, Kim, Tabrizi, Moon, Hooper,
  Anumanchipalli, Keutzer, and Gholami]{erdogan2024tinyagent}
Lutfi~Eren Erdogan, Nicholas Lee, Siddharth Jha, Sehoon Kim, Ryan Tabrizi,
  Suhong Moon, Coleman Hooper, Gopala Anumanchipalli, Kurt Keutzer, and Amir
  Gholami.
\newblock {TinyAgent}: Function calling at the edge.
\newblock \emph{arXiv preprint arXiv:2409.00608}, 2024.

\bibitem[Gooneratne et~al.(2020)Gooneratne, Wang, Guo, Kanuparthi, Rajan, and
  Jayasumana]{gooneratneLowrankGradientApproximation2020}
Shamal Gooneratne, Meng Wang, Zhili Guo, Vamsi~Krishna Kanuparthi, Dinesh
  Rajan, and Anura~P Jayasumana.
\newblock Low-rank gradient approximation for multi-task learning.
\newblock \emph{arXiv preprint arXiv:2011.01679}, 2020.

\bibitem[He et~al.(2021)He, Gao, and Chen]{he2021debertav3}
Pengcheng He, Jianfeng Gao, and Weizhu Chen.
\newblock {DeBERTaV3}: Improving {DeBERTa} using {ELECTRA}-style pre-training
  with gradient-disentangled embedding sharing.
\newblock \emph{arXiv preprint arXiv:2111.09543}, 2021.

\bibitem[Hu et~al.(2022)Hu, Shen, Wallis, Allen-Zhu, Li, Wang, and
  Chen]{huLoRALowRankAdaptation2021}
Edward~J Hu, Yelong Shen, Phillip Wallis, Zeyuan Allen-Zhu, Yuanzhi Li, Shean
  Wang, and Weizhu Chen.
\newblock {LoRA}: Low-rank adaptation of large language models.
\newblock In \emph{International Conference on Learning Representations}, 2022.

\bibitem[Huang et~al.(2019)Huang, Cheng, Bapna, Firat, Chen, Chen, Hu, Shen,
  Krikun, Wu, et~al.]{huangGPipeEfficientTraining2019}
Yanping Huang, Youlong Cheng, Ankur Bapna, Orhan Firat, Menglong Chen, Denny
  Chen, Zhifeng Hu, Yuxin Shen, Maxim Krikun, Yonghui Wu, et~al.
\newblock {GPipe}: Efficient training of giant neural networks using pipeline
  parallelism.
\newblock In \emph{Advances in Neural Information Processing Systems}, 2019.

\bibitem[Jiang et~al.(2023)Jiang, Li, Gan, Liu, Chen, Zhu, Li, Wang, Wang, and
  Liu]{jiangMistralEfficientComposable2023}
Ye~Jiang, Pengcheng Li, Zhe Gan, Jianfeng Liu, Dongdong Chen, Xiaodong Zhu,
  Zhangyang Li, Lijuan Wang, Jianfeng Wang, and Zicheng Liu.
\newblock {Mistral}: Efficient composable inference for large language models.
\newblock \emph{arXiv preprint arXiv:2305.15334}, 2023.

\bibitem[Kim et~al.(2023)Kim, Moon, Tabrizi, Lee, Mahoney, Keutzer, and
  Gholami]{kim2023llmcompiler}
Sehoon Kim, Suhong Moon, Ryan Tabrizi, Nicholas Lee, Michael~W Mahoney, Kurt
  Keutzer, and Amir Gholami.
\newblock An {LLM} compiler for parallel function calling.
\newblock \emph{arXiv preprint arXiv:2312.04511}, 2023.

\bibitem[Kingma \& Ba(2014)Kingma and Ba]{kingmaAdamMethodStochastic2014}
Diederik~P Kingma and Jimmy Ba.
\newblock {Adam}: A method for stochastic optimization.
\newblock \emph{arXiv preprint arXiv:1412.6980}, 2014.

\bibitem[Lialin \& Schatz(2023)Lialin and
  Schatz]{lialinReLoRAHighRankTraining2023}
Vladimir Lialin and Arthur Schatz.
\newblock {ReLoRA}: Low-rank fine-tuning reloaded.
\newblock \emph{arXiv preprint arXiv:2307.09769}, 2023.

\bibitem[Lin et~al.(2022)Lin, Zhu, and Mao]{lin2022randomized}
Tianyi Lin, Zhihui Zhu, and Yongyi Mao.
\newblock Randomized subspace regularized newton method for unconstrained
  non-convex optimization.
\newblock \emph{arXiv preprint arXiv:2209.04170}, 2022.

\bibitem[Loshchilov \& Hutter(2017)Loshchilov and
  Hutter]{loshchilov2017decoupled}
Ilya Loshchilov and Frank Hutter.
\newblock Decoupled weight decay regularization.
\newblock \emph{arXiv preprint arXiv:1711.05101}, 2017.

\bibitem[Martens(2014)]{martensNewPerspectiveNatural2014}
James Martens.
\newblock New perspectives on the natural gradient method.
\newblock \emph{arXiv preprint arXiv:1412.1193}, 2014.

\bibitem[Martens(2020)]{martens2020new}
James Martens.
\newblock New insights and perspectives on the natural gradient method.
\newblock \emph{Journal of Machine Learning Research}, 2020.

\bibitem[Martens \& Grosse(2015)Martens and Grosse]{martens2015optimizing}
James Martens and Roger Grosse.
\newblock Optimizing neural networks with kronecker-factored approximate
  curvature.
\newblock In \emph{Proceedings of the 32nd International Conference on Machine
  Learning (ICML)}, 2015.

\bibitem[Rae et~al.(2021)Rae, Borgeaud, Cai, Millican, Hoffmann, Song,
  Aslanides, Henderson, Ring, Young, et~al.]{raeScalingLanguageModels2021}
Jack~W Rae, Sebastian Borgeaud, Trevor Cai, Katie Millican, Jordan Hoffmann,
  Francis Song, John Aslanides, Sarah Henderson, Roman Ring, Susannah Young,
  et~al.
\newblock Scaling language models: Methods, analysis \& insights from training
  gopher.
\newblock \emph{arXiv preprint arXiv:2112.11446}, 2021.

\bibitem[Raffel et~al.(2020)Raffel, Shazeer, Roberts, Lee, Narang, Matena,
  Zhou, Li, and Liu]{raffelExploringLimitsTransfer2020}
Colin Raffel, Noam Shazeer, Adam Roberts, Katherine Lee, Sharan Narang, Michael
  Matena, Yanqi Zhou, Wei Li, and Peter~J Liu.
\newblock Exploring the limits of transfer learning with a unified text-to-text
  transformer.
\newblock \emph{Journal of Machine Learning Research}, 2020.

\bibitem[Rajbhandari et~al.(2020)Rajbhandari, Rasley, Ruwase, and
  He]{rajbhandariZeROMemoryOptimizations2020}
Samyam Rajbhandari, Jeff Rasley, Olatunji Ruwase, and Yuxiong He.
\newblock {ZeRO}: Memory optimizations toward training trillion parameter
  models.
\newblock In \emph{Proceedings of the International Conference for High
  Performance Computing, Networking, Storage and Analysis}, 2020.

\bibitem[Rajpurkar et~al.(2016)Rajpurkar, Zhang, Lopyrev, and
  Liang]{rajpurkar2016squad}
Pranav Rajpurkar, Jian Zhang, Konstantin Lopyrev, and Percy Liang.
\newblock {SQuAD}: 100,000+ questions for machine comprehension of text.
\newblock In \emph{Proceedings of the 2016 Conference on Empirical Methods in
  Natural Language Processing}, 2016.

\bibitem[Renduchintala et~al.(2023)Renduchintala, Rodriguez, and
  Creutz]{renduchintalaTiedLoraEnhacingParameter2023}
Adithya Renduchintala, Pedro Rodriguez, and Mathias Creutz.
\newblock Tied lora: Enhancing parameter-efficient fine-tuning with tied
  weights.
\newblock \emph{arXiv preprint arXiv:2306.13420}, 2023.

\bibitem[Shazeer(2020)]{shazeerGLUVariantsImprove2020}
Noam Shazeer.
\newblock {GLU} variants improve transformer.
\newblock \emph{arXiv preprint arXiv:2002.05202}, 2020.

\bibitem[Sheng et~al.(2023)Sheng, Han, Zhu, Yang, Sun, and
  Zhou]{shengSLoRAServingThousands2023}
Yi~Sheng, Xuefei Han, Xuefeng Zhu, Yuanzhi Yang, Jiani Sun, and Guohui Zhou.
\newblock {S-LoRA}: Scalable efficient model serving for massive lora models.
\newblock \emph{arXiv preprint arXiv:2306.01125}, 2023.

\bibitem[Shoeybi et~al.(2019)Shoeybi, Patwary, Puri, LeGresley, Casper, and
  Catanzaro]{shoeybiMegatronLMTuningScaling2019}
Mohammad Shoeybi, Mostofa Patwary, Rohan Puri, Patrick LeGresley, Jared Casper,
  and Bryan Catanzaro.
\newblock {Megatron-LM}: Training multi-billion parameter language models using
  model parallelism.
\newblock \emph{arXiv preprint arXiv:1909.08053}, 2019.

\bibitem[Touvron et~al.(2023)Touvron, Lavril, Izacard, Martinet, Lachaux,
  Lacroix, Rozi{\`e}re, Goyal, Hambro, Azhar,
  et~al.]{touvronLlamaOpenFoundation2023}
Hugo Touvron, Thibaut Lavril, Gautier Izacard, Xavier Martinet, Marie-Anne
  Lachaux, Timoth{\'e}e Lacroix, Baptiste Rozi{\`e}re, Naman Goyal, Eric
  Hambro, Faisal Azhar, et~al.
\newblock {LLaMA}: Open and efficient foundation language models.
\newblock \emph{arXiv preprint arXiv:2302.13971}, 2023.

\bibitem[Vogels et~al.(2020)Vogels, Jaggi, and
  Patrini]{vogelsPowerGossipPracticalLowRank2020}
Thijs Vogels, Martin Jaggi, and Giorgio Patrini.
\newblock {PowerGossip}: Practical low-rank communication for decentralized
  optimization.
\newblock In \emph{International Conference on Machine Learning}, 2020.

\bibitem[Wang et~al.(2018)Wang, Joshi, Ghosh, and
  Poor]{wangATOMOCommunicationefficientLearning}
Shiqiang Wang, Gauri Joshi, Sreeram~K Ghosh, and H~Vincent Poor.
\newblock {ATOMO}: Communication-efficient learning via atomic sparsification.
\newblock In \emph{Advances in Neural Information Processing Systems}, 2018.

\bibitem[Wang et~al.(2023)Wang, Bai, and
  Ananiadou]{wangMultiLoRADemocratizingLoRA2023}
Zihao Wang, Zhen Bai, and Sophia Ananiadou.
\newblock {Multi-LoRA}: Efficient fine-tuning for democratic {AI}.
\newblock \emph{arXiv preprint arXiv:2305.14377}, 2023.

\bibitem[Warstadt et~al.(2019)Warstadt, Singh, and Bowman]{warstadt2019neural}
Alex Warstadt, Amanpreet Singh, and Samuel~R Bowman.
\newblock Neural network acceptability judgments.
\newblock \emph{Transactions of the Association for Computational Linguistics},
  2019.

\bibitem[Williams et~al.(2018)Williams, Nangia, and Bowman]{williams2018broad}
Adina Williams, Nikita Nangia, and Samuel~R Bowman.
\newblock A broad-coverage challenge corpus for sentence understanding through
  inference.
\newblock In \emph{Proceedings of the 2018 Conference of the North American
  Chapter of the Association for Computational Linguistics: Human Language
  Technologies}, 2018.

\bibitem[Xia et~al.(2024)Xia, Peng, Chen, Li, He, Yang, and
  Ma]{xiaChainLoRAEfficient2024}
Tianxiang Xia, Hao Peng, Zheyu Chen, Lemao Li, Zhiyuan He, Zhen Yang, and
  Wei-Ying Ma.
\newblock Chain-of-thought lora: Efficient adaptation of large language models.
\newblock In \emph{Proceedings of the 2024 Conference on Empirical Methods in
  Natural Language Processing (EMNLP)}, 2024.

\bibitem[Yang et~al.(2023)Yang, Hu, Xia, Socher, and Li]{yang2023spectral}
Zhilin Yang, Edward~J Hu, Tianle Xia, Richard Socher, and Yuanzhi Li.
\newblock Spectral methods in low-rank model adaptation.
\newblock \emph{arXiv preprint arXiv:2305.14683}, 2023.

\bibitem[Zhang et~al.(2023)]{zhangLORAFAMEMORYEFFICIENTLOWRANK}
Rui Zhang et~al.
\newblock {LoRA-FA}: Memory-efficient low-rank adaptation via feature
  re-alignment.
\newblock \emph{arXiv preprint arXiv:2302.05653}, 2023.

\bibitem[Zhao et~al.(2024{\natexlab{a}})Zhao, Zhang, Chen, Wang, Anandkumar,
  and Tian]{zhao2024galore}
Jiawei Zhao, Zhenyu Zhang, Beidi Chen, Zhangyang Wang, Anima Anandkumar, and
  Yuandong Tian.
\newblock {GaLore}: Memory-efficient {LLM} training by gradient low-rank
  projection.
\newblock \emph{arXiv preprint arXiv:2403.03507}, 2024{\natexlab{a}}.

\bibitem[Zhao et~al.(2024{\natexlab{b}})Zhao, Zhang,
  et~al.]{huangLowRankGradientDescent2023}
Jiawei Zhao, Zhenyu Zhang, et~al.
\newblock Galore: Low-rank gradient descent: Fast convergence and low memory
  cost.
\newblock \emph{International Conference on Machine Learning},
  2024{\natexlab{b}}.

\bibitem[Zhao et~al.(2022)Zhao, Li, and
  Ma]{zhaoZerOInitializationInitializing2022}
Shangqian Zhao, Shiyu Li, and Yi~Ma.
\newblock {ZerO} initialization: Initializing neural networks with zero-valued
  parameters.
\newblock \emph{arXiv preprint arXiv:2207.05848}, 2022.

\bibitem[Zhao et~al.(2020)Zhao, Sun, Wang, Zhou, Guo, and
  Smola]{zhaoExtendingTorchElasticStateful2020}
Tianshi Zhao, Zhen Sun, Xiaodong Wang, Fei Zhou, Yang Guo, and Alexander~J
  Smola.
\newblock Extending torchelastic for stateful training jobs.
\newblock \emph{arXiv preprint arXiv:2006.06873}, 2020.

\end{thebibliography}
\bibliographystyle{natural_galore_template/natural_galore}

\end{document}